\documentclass{article}
\usepackage{spconf}
\usepackage{cite}
\usepackage{amsmath,amssymb,amsfonts}
\usepackage{algorithmic}
\usepackage{graphicx}
\usepackage{textcomp}
\usepackage{xcolor}
\usepackage{booktabs}
\usepackage{multirow}
\usepackage{xspace}
\usepackage{url}
\usepackage{hyperref}
\usepackage{caption}
\makeatletter
\renewcommand\section{\@startsection{section}{1}{\z@}%
  {2.0ex plus 0.2ex minus 0.1ex}
  {1.0ex plus 0.2ex}
  {\relax}
}

\renewcommand\subsection{\@startsection{subsection}{2}{\z@}%
  {1.0ex plus 0.2ex minus 0.1ex}
  {0.5ex plus 0.1ex}
  {\relax}%
}

\renewcommand\subsubsection{\@startsection{subsubsection}{3}{\z@}%
  {1ex plus 0.2ex minus 0.1ex}
  {0.3ex plus 0.1ex}
  {\relax}%
}

\renewcommand\thebibliography[1]{%
  \section{References}%
  \list{[\arabic{enumi}]}{%
    \settowidth\labelwidth{[#1]}%
    \leftmargin\labelwidth
    \advance\leftmargin\labelsep
    \usecounter{enumi}%
    \itemsep=2pt
    \parsep=0pt
    \topsep=0pt
  }%
  \sloppy
}

\makeatother

\newcommand{\etal}{\textit{et al.}\xspace}

\title{CPDDNet: Color-Polarization Denoising and Demosaicking Network}
\name{Qihang Zhang$^{1,2}$ \qquad Yusuke Monno$^{1}$ \qquad Masayuki Tanaka$^{1}$ \qquad Masatoshi Okutomi$^{1}$}
\address{
$^{1}$ Institute of Science Tokyo \qquad \qquad
$^{2}$ The Chinese University of Hong Kong, Shenzhen \\
}

\begin{document}
\maketitle
\begin{abstract}
Color-polarization imaging using a color-polarization filter array~(CPFA) sensor captures both texture~(color intensity) and physical~(polarization) information of the scene in a single shot, enabling various applications in computer vision. 
However, the raw mosaic output from a CPFA sensor often suffers from severe noise and resolution loss, especially under low-light conditions. 
Existing methods generally focus on either denoising or demosaicking tasks, failing to capture the coupling between them and neglecting shared low-level features.
In this paper, we propose a color-polarization denoising and demosaicking network~(CPDDNet), which is a joint framework that performs noise removal and CPFA interpolation using a feature fusion module that retains the features from the CPFA raw data at both the denoising and the demosaicking stages. 
Experimental results demonstrate that CPDDNet significantly enhances image quality and polarization parameter accuracy, outperforming existing approaches on a real dataset. The source code of CPDDNet will be publicly available upon the paper acceptance.
\end{abstract}

\section{Introduction}
\label{sec:intro}

\begin{figure}[htb]
\begin{minipage}[b]{1.0\linewidth}
  \centering
  \centerline{\includegraphics[width=8.8cm]{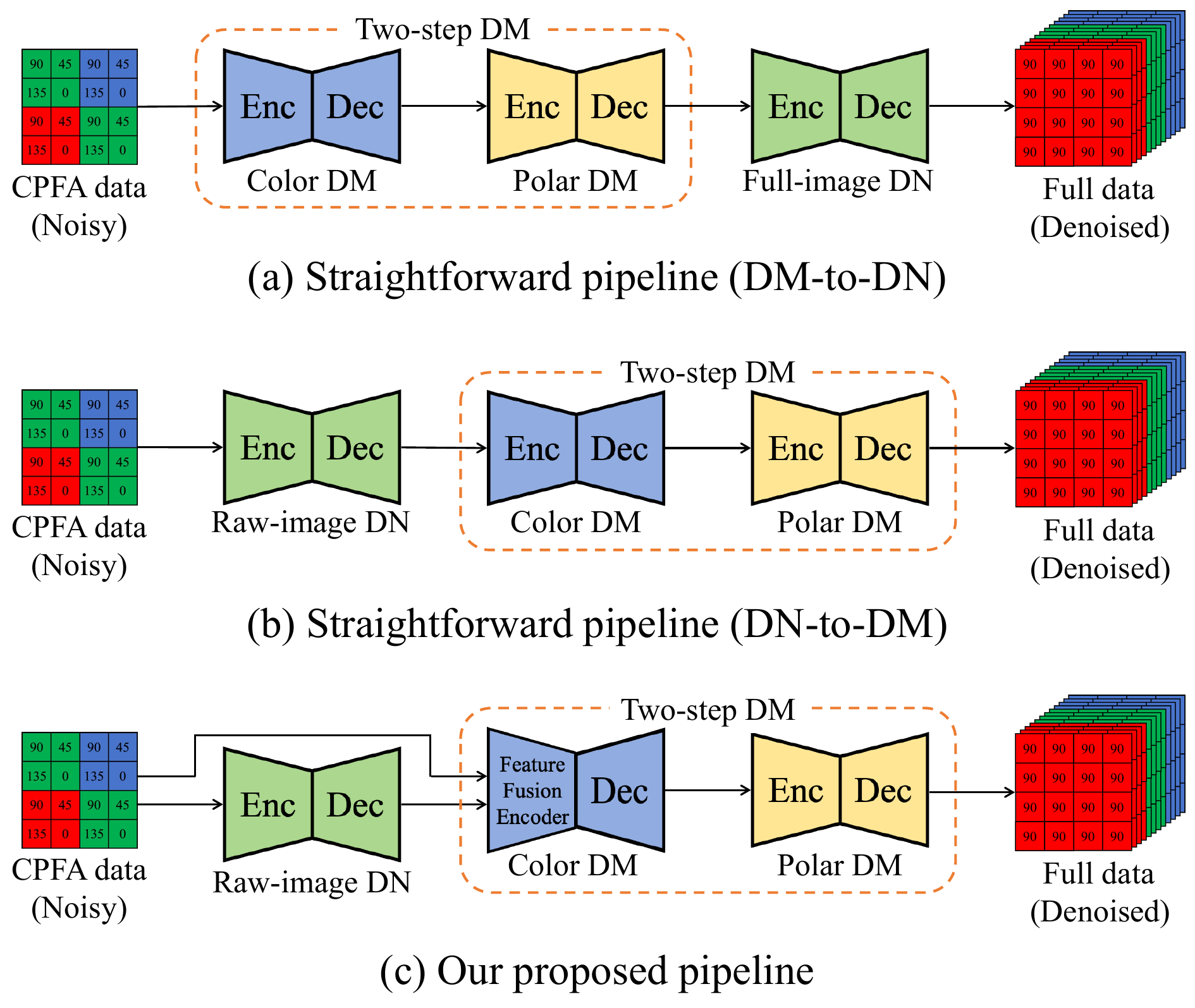}} \vspace{-3mm}
\end{minipage}
\caption{The overviews of different pipelines. The straightforward pipelines sequentially apply demosaicking (DM) and denoising (DN) as DM-to-DN~(a) or DN-to-DM~(b). While our proposed pipeline~(c) is based on the DN-to-DM pipeline, we newly introduce a feature fusion encoder to retain the information from the raw CPFA data at both the DN and the DM stages, reducing the information loss in the whole pipeline.}
\label{fig:pipeline}
\end{figure}

Color-polarization imaging captures both color texture and polarization information of the scene
and has various computer vision applications, such as material segmentation~\cite{liang2022multimodal} and 3D reconstruction~\cite{cui2017polarimetric}.
A division-of-focal-plane polarimeter equips a color–polarization filter array~(CPFA, see Fig.~\ref{fig:pipeline}) on an image sensor, enabling the acquisition of four polarization-orientation data ($0^\circ$, $45^\circ$, $90^\circ$, and $135^\circ$) in one shot~\cite{8354948}. However, since CPFA consists of a mosaic of 12-channel data (four polarization orientations and three color channels), an interpolation process of missing information called demosaicking~(DM) is required to obtain full 12-channel color-polarization data. In addition, CPFA data suffer from noise in some scenarios, such as fast~(short-exposure) and low-light imaging, necessitating denoising~(DN) at the same time.

Most existing studies focus on either DM or DN in isolation, while little attention has been paid to designing a unified pipeline for CPFA sensors. As a result, current solutions rely on a simple sequential combination of DM and DN, arranged sequentially as DM-to-DN or DN-to-DM. 
However, these separated approaches overlook the strong coupling between DM and DN, which not only prevents feature sharing but also propagates errors. In particular, DM-to-DN tends to propagate noise during DM,whereas DN-to-DM often oversmooths signal-dependent structures during DN.
Although CPFA-specific DM or DN methods have been developed~(e.g.,~\cite{9583277, zhou2025pidsr, Liang:22, 10203642}), they still lack a holistic design for CPFA imaging with DM and DN as core components and thus perform poorly in high-noise conditions.

In this paper, we propose a color-polarization denoising and demosaicking network, namely CPDDNet. Figure~\ref{fig:pipeline} illustrates the differences between the straightforward pipelines (DM-to-DN and DN-to-DM) and the pipeline of our CPDDNet, where we consider the two-step approach of~\cite{nguyen2022two} for DM. Although our pipeline is based on the DN-to-DM pipeline, we newly introduce a feature fusion encoder to DM that fuses the features from the noisy raw CPFA image with those from the denoised CPFA image. This retains the information from the raw CPFA data at both the DN and the DM stages, thereby reducing information loss in the whole pipeline. We evaluate our proposed CPDDNet on a recently published real dataset of~\cite{11084558} and demonstrate its superior performance. The main contributions of this study are summarized as follows.

\begin{itemize}
    \item We propose CPDDNet, which is the first network architecture combining DN and DM for CPFA sensors, to the best of our knowledge.
    \item We design a unified DN-to-DM pipeline equipped with a feature fusion encoder, enabling information bridging between the DN and the DM stages.
    \item We experimentally validate that our CPDDNet achieves state-of-the-art performance and outperforms existing approaches both numerically and visually. The source code is publicly available at our project page\footnote{\thanks{\url{http://www.ok.sc.e.titech.ac.jp/res/PolarDem/CPDDNet/}}}.
\end{itemize}

\section{Related Work}
\label{sec:prior}

\begin{figure*}[t!]
\centering
\includegraphics[width=\textwidth]{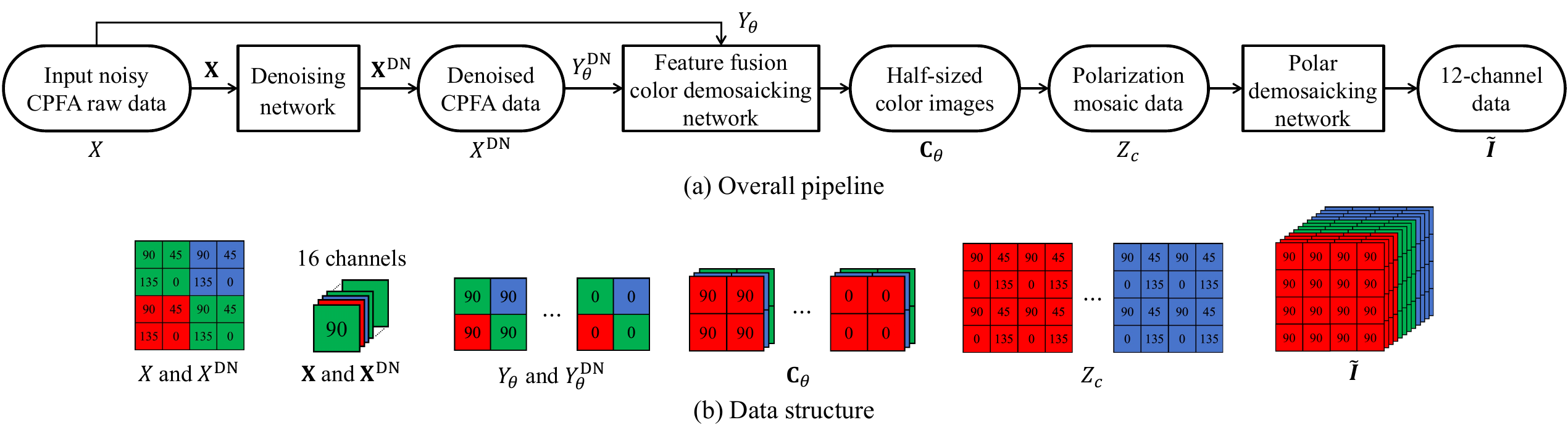} \\ \vspace{-4mm}
\caption{The overall pipeline~(a) and data structure~(b) of our proposed method.}
\label{fig:unet}
\end{figure*}

\begin{figure*}[t!]
\centering
\includegraphics[width=1\textwidth]{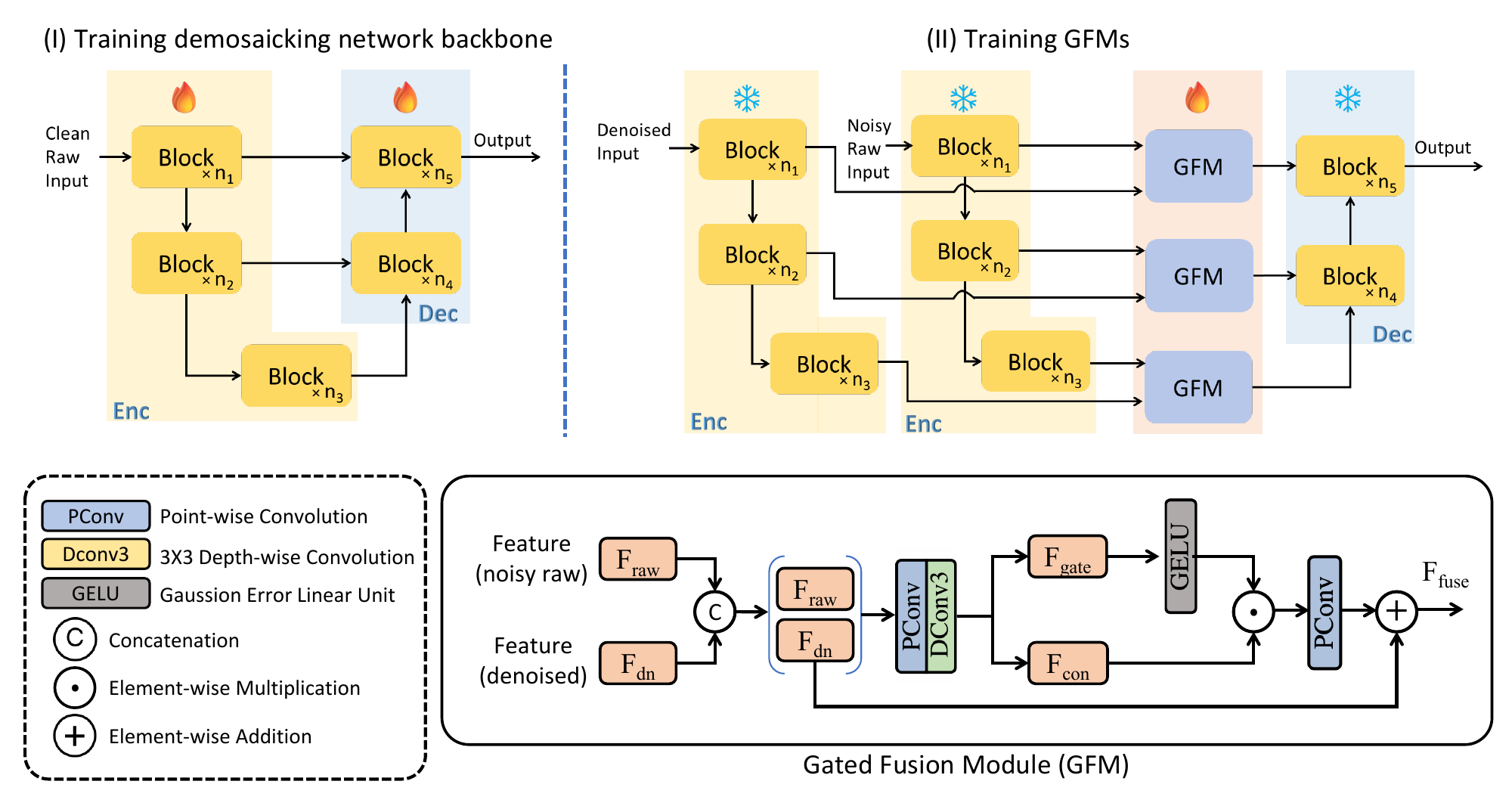}\\ \vspace{-3mm}
\caption{Two-stage training strategy and the detailed structure of our color DM network with GFMs. In Stage~I, the encoder and the decoder are trained using clean (noise-free) CPFA raw data to learn only the DM task. In Stage~II, the GFMs are trained while keeping the encoder and the decoder parameters frozen to address the joint task of DN and DM. The fire icon indicates components being trained in the current stage, while the snowflake icon indicates frozen parameters.}
\label{fig:netpipeline}
\end{figure*}

While CFPA-specific DN~\cite{Liang:22,10203642} and DM~\cite{9583277,nguyen2022two,zhou2025pidsr,su2025universal,wang2025swincpd} methods have been developed in both signal-processing-based~\cite{Liang:22,9583277,su2025universal} and deep-learning-based~\cite{10203642,nguyen2022two,zhou2025pidsr,wang2025swincpd} manners, only a few studies focus on jointly addressing DN and DM for CPFA sensors. A recent study of~\cite{11084558} addressed the lack of an evaluation dataset for this joint task and constructed a new polarization image dataset containing 40 scenes with three noise-level conditions. This study also constructed a baseline signal-processing-based method, which adopts the DN-to-DM pipeline and demonstrates its effectiveness compared with the DM-to-DN pipeline. While this baseline method effectively combines the pseudo four-channel color denoising (PFCD) of~\cite{7351714} for DN and the intensity-guided residual interpolation (IGRI) of~\cite{9583277} for DM, a method utilizing powerful deep learning techniques is expected to outperform this signal-processing-based baseline method.   

To our knowledge, only Li \etal~\cite{10304085} explicitly perform joint DN and DM, targeting monochrome long-wave infrared division-of-focal-plane
polarization images with mixed-noise estimation. However, the task is not designed for CPFA sensors, and the approach is essentially a simple stacking of DN and DM modules without specific designs for the joint task.

When joint DN and DM methods are absent, the sequential pipelines of DN-to-DM and DM-to-DN are common options. However, these straightforward pipelines are likely to produce suboptimal results, since noise is typically propagated during DM, and image structures are often oversmoothed during DN.  

With the above backgrounds, in this paper, we propose CPDDNet, which is the first network architecture combining DN and DM for CPFA sensors, to our knowledge. Following the observations and the results in~\cite{11084558}, we adopt the DN-to-DM pipeline as the base and newly introduce a feature fusion module to bridge the information between DN and DM, reducing the information loss in the whole pipeline.

\section{Proposed Method}
\label{sec:method}

\subsection{Problem Formulation}

In this study, we consider the $4\times4$ pattern of Sony's CPFA sensors~\cite{8354948}, as shown in Fig.~\ref{fig:pipeline}, and aim to reconstruct the full noise-free 12-channel color-polarization data described as
\begin{equation}\label{eq:cube_def}
\mathbf{I} = \{ I_{\theta}^c \} \in \mathbb{R}^{H\times W \times (4\times3)},
\end{equation}
where $H \times W$ represents image height and width, $\theta \in \{0^\circ,45^\circ,90^\circ,135^\circ\}$ is a polarization orientation, and $c \in \{R,G,B\}$ is a color channel.
The input noisy CPFA raw data $X \in \mathbb{R}^{H\times W}$ is expressed as
\begin{align}\label{eq:sampling}
X = \sum_{k=1}^{12} M_k\odot I_k + N,
\end{align}
where $I_k \in \mathbb{R}^{H\times W}$ is the $k$-th channel image of $\mathbf{I}$, $M_k\in\{0,1\}^{H\times W}$ is a binary mask for each channel satisfying $\sum_{k} M_k(i,j)=1$, representing that only one channel among 12 channels is measured at each pixel $(i,j)$ in the CPFA data, $\odot$ denotes the Hadamard product, and $N \in \mathbb{R}^{H\times W}$ represents measurement noise. The joint task of polarization DN and DM is to reconstruct $\mathbf{I}$ from $X$.

\subsection{Overall Pipeline}  

Figure~\ref{fig:unet} shows the overall pipeline and data structure of our proposed method, which consists of three networks, as explained below. 

1) CPFA-domain DN: We first rearrange the $4 \times 4$ mosaic CPFA pattern into a 16-channel tensor as $\mathbf{X} \in \mathbb{R}^{\frac{H}{4} \times \frac{W}{4} \times 16}$. Then, we apply a U-Net architecture~\cite{ronneberger2015u} to that tensor to remove noise while preserving signal-dependent structures. This stage follows the polarization-aware DN paradigm of~\cite{10203642} and produces a denoised tensor \(\mathbf{X}^{\text{DN}}\). This is returned to the mosaic pattern, which is expressed as $X^{\text{DN}} \in \mathbb{R}^{H\times W}$.

2) Bayer-domain color DM: In the DM stage, we follow the two-step approach of~\cite{nguyen2022two}, which first applies color DM. From the denoised CPFA data $X^{\text{DN}}$, Bayer-patterned images~\cite{bayer1976color} for each polarization orientation are constructed by $2 \times 2$ sub-sampling as $Y^{\text{DN}}_{\theta} \in \mathbb{R}^{\frac{H}{2} \times \frac{W}{2}}$. While the straightforward DN-to-DM pipeline only adopts the denoised input, we add the noisy Bayer-patterned images $Y_{\theta}$ constructed from the original CFPA raw data $X$ as the input for color DM. The multi-scale features extracted from the two inputs are fused by a gated fusion module (GFM), inspired by the decouple-and-feedback fusion~\cite{10204662}, adaptively combining high-frequency details from the raw input and robust low-frequency structures from the denoised input. The details of GFM are explained in the next subsection. The outputs of color DM are half-sized three-channel color images for each polarization orientation, which are expressed as $\mathbf{C}_{\theta} \in \mathbb{R}^{\frac{H}{2} \times \frac{W}{2} \times 3}$.

3) Polar DM: The final step is to apply polar DM. From the set of half-sized three-channel color images for all polarization orientations $\{\mathbf{C}_{0}, \mathbf{C}_{45}, \mathbf{C}_{90}, \mathbf{C}_{135}\}$, the polarization mosaic data for each color channel is constructed by pixel rearrangement, which is expressed as $Z_c \in \mathbb{R}^{H\times W}$. This is fed into the polar DM, which produces the full polarization image for each color channel, denoted by $\mathbf{P}_c \in \mathbb{R}^{H\times W \times 4}$. We follow the network architecture of~\cite{nguyen2022two} in this part. Finally, the reconstructed full 12-channel color-polarization data $\tilde{\mathbf{I}}$ is formed by stacking all the outputs as $\tilde{\mathbf{I}} = Concat[\mathbf{P}_R, \mathbf{P}_G, \mathbf{P}_B]$, where $Concat$ represents the concatenation along the channel dimension.

\subsection{Gated Fusion Module (GFM)}

The right side of Fig.~\ref{fig:netpipeline} shows the architecture of our color DM network with GFMs. We adopt a U-Net architecture as the backbone~\cite{ronneberger2015u} and introduce dual-encoder branches. 
Our GFM adaptively fuses the features from the dual-encoder branches so that high-frequency features from the raw input are retained while noise-suppressed features from the denoised input represent the base image structures. This design follows the spirit of the decouple-and-feedback fusion~\cite{10204662}, and employs efficient point-wise and depth-wise convolutions widely used in modern restoration networks \cite{Zamir2021Restormer}.
At stage~$l$, the features from the raw and the denoised branches are first mixed and then split channel-wise into a gating tensor $F_{gate}$ and a content tensor $F_{con}$. The gate, after a Gaussian error linear unit~(GELU) activation, modulates the content via element-wise multiplication, and the result is projected and residually combined with the features from the denoised branch. The operations at the $l$-th ($l \in \{1,2,...,L\}$) stage are described as
\begin{align}
    F^{l}_{gate}, F^{l}_{con} &= \text{DConv3}(\text{PConv}(Concat[F^{l}_{raw}, F^{l}_{dn}])), \label{eq:gfm1} \\
    F^{l}_{fuse} &= \text{PConv}(F^{l}_{con} \odot \text{GELU}(F^{l}_{gate})) + F^{l}_{dn}, \label{eq:gfm2}
\end{align}
where $F^{l}_{raw}$ and $F^{l}_{dn}$ are the features from the raw and the denoised encoder branches, respectively. PConv and DConv3 represent a point-wise $1\times1$ convolution and a depth-wise convolution with a $3 \times 3$ kernel, respectively. $\odot$ denotes the Hadamard product and $F^{l}_{\text{fuse}}$ is the fused output forwarded to the decoder. 

\subsection{Loss Functions}

For the DN network, we take an L1 loss between the denoised output $\mathbf{X}^{\text{DN}}$ and the corresponding ground truth $\mathbf{X}^{\text{GT}}$ as
\begin{equation}
\mathcal{L}_{DN} = 
\frac{1}{HW}
\left\|\mathbf{X}^{\text{DN}} - \mathbf{X}^{\text{GT}}\right\|_{1}.
\label{eq:loss_d}
\end{equation}

For the color DM and the polar DM networks, we employ the same loss function as used in TCPDNet~\cite{nguyen2022two}. 
In short, TCPDNet uses a sub-sampled RGB reconstruction term $\mathcal{L}_{C}$ and a full-resolution color-polarization reconstruction term (in YCbCr) $\mathcal{L}_{CP}^{YCbCr}$. The overall DM loss is described as
\begin{equation}
\mathcal{L_{DM}} = \mathcal{L}_{C} + \alpha \mathcal{L}_{CP}^{YCbCr},
\label{eq:loss_total}
\end{equation}
where $\alpha$ is fixed to $4$, following~\cite{nguyen2022two}. Each term is described as
\begin{align}
\mathcal{L}_{C} &= \frac{1}{3HW}
\sum_{\theta}\left\|\mathbf{C}_{\theta} - \mathbf{C}_{\theta}^{\text{GT}}\right\|_{1},\\
\mathcal{L}_{CP}^{YCbCr} &= \frac{1}{12HW}
\left\|YCbYr(\tilde{\mathbf{I}}) - YCbCr(\mathbf{I}^{\text{GT}})\right\|_{1}.
\end{align}
where $\mathbf{C}_{\theta}^{\text{GT}}$ and $\mathbf{I}^{\text{GT}}$ are the corresponding ground truths, and $YCbCr$ represents the transformation from the RGB space to the YCbCr space.
We refer to the paper~\cite{nguyen2022two} for more detailed information.

\subsection{Training Strategy}

Stage I: Separated DN and DM training. The first stage trains the DN and the DM network backbones separately. For DN, we train the DN network using the loss function of Eq.~(\ref{eq:loss_d}). For DM, we train the color DM and the polar DM networks using the loss function of Eq.~(\ref{eq:loss_total}), where we remove the dual-encoder branches and GFMs from the color DM network and use clean (noise-free) CPFA raw data as the input for DM to learn only the DM task, as shown in Fig.~\ref{fig:netpipeline}. 

Stage-II: GFM training. We freeze all network parameters obtained in Stage I for the DN and the DM backbones. Then, we add our proposed dual-encoder branches and train GFMs for the DM stage using the same loss function, where we use the noisy CPFA raw data and its denoised version as the network inputs, as shown in Fig.~\ref{fig:netpipeline}.

\begin{table*}[t!]
\centering
\normalsize
\caption{Quantitative comparison on the dataset~\cite{11084558}, with the best performance highlighted in bold.}
\label{tab:tokyo_results}
\renewcommand{\arraystretch}{1.2}
\setlength{\tabcolsep}{4pt}
\begin{tabular}{llccccccc}
\toprule
\textbf{Category} & \textbf{Method} & 
\multicolumn{6}{c}{\textbf{PSNR$\uparrow$/SSIM$\uparrow$}} & 
\textbf{MAE$\downarrow$} \\
\cmidrule(lr){3-8} \cmidrule(lr){9-9}
& & $I_0$ & $I_{45}$ & $I_{90}$ & $I_{135}$ & $S_0$ & DoP & AoP \\
\midrule
\multirow{2}{*}{DM Only} 
& IGRI-2 & 29.16/0.675 & 29.04/0.673 & 29.07/0.672 & 29.01/0.670 & 31.57/0.742 & 20.10/0.251 & 46.98 \\
& TCPDNet & 30.10/0.660 & 29.37/0.629 & 29.97/0.658 & 29.60/0.633 & 30.31/0.668 & 26.03/0.433 & 46.83 \\
\midrule
\multirow{2}{*}{DM-to-DN}
& IGRI-2 $\rightarrow$ BM3D & 32.40/0.886 & 32.17/0.882 & 32.28/0.883 & 32.17/0.881 & 35.87/0.933 & 22.97/0.446 & 45.05 \\
& TCPDNet $\rightarrow$ U-Net  & 33.12/0.805 & 32.59/0.787 & 32.77/0.793 & 32.77/0.788 & 33.42/0.810 & 31.04/0.619 & 46.83 \\
\midrule
\multirow{2}{*}{DN-to-DM} 
& PFCD $\rightarrow$ IGRI-2   & 34.29/0.935 & 34.06/0.932 & 34.35/0.935 & 34.20/0.933 & 35.59/0.947 & 29.59/0.686 & 41.84 \\
& U-Net $\rightarrow$ TCPDNet    & 36.15/0.939 & 35.76/0.934 & 36.08/0.939 & 36.01/0.935 & 36.56/0.942 & 34.21/0.805 & 42.24 \\
\midrule
\multirow{1}{*}{Ours} 
& CPDDNet & \textbf{36.31/0.943} & \textbf{36.03/0.939} & \textbf{36.31/0.942} & \textbf{36.29/0.940} & \textbf{36.78/0.946} & \textbf{34.50/0.816} & \textbf{41.47} \\
\bottomrule
\end{tabular}
\end{table*}

\begin{figure*}[t!]
\centering
\vspace{3mm}
\includegraphics[width=\textwidth]{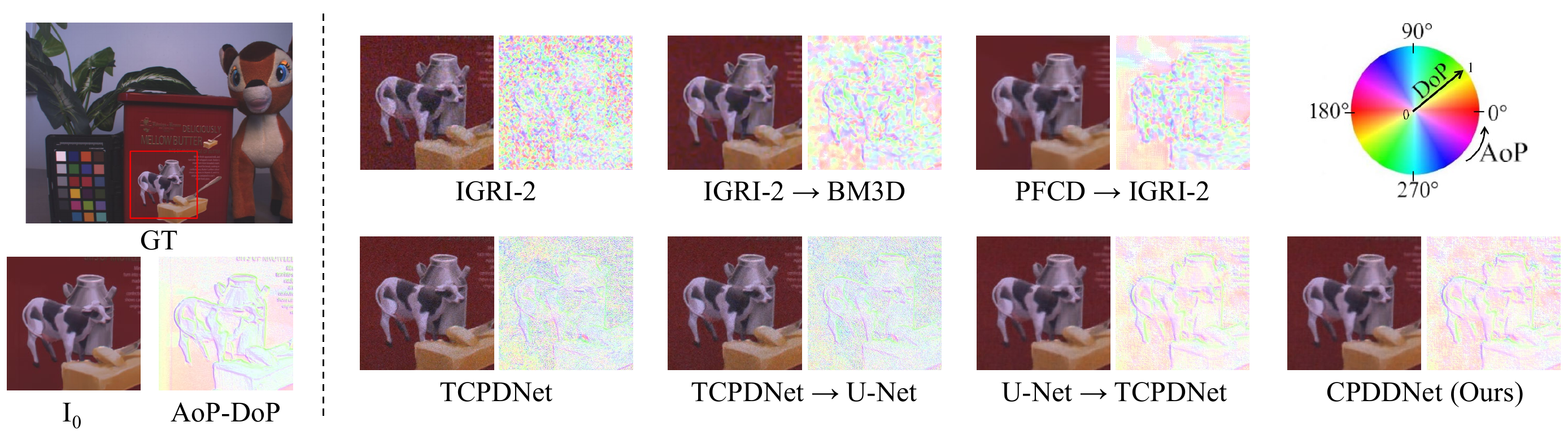}\\ \vspace{-3mm}
\caption{Visual comparisons on one of the test scenes. The red box highlights the region for zoomed-in inspection. Additional qualitative comparisons can be found in the supplementary material.}
\label{fig:vis}
\end{figure*}

\section{Experimental Results}

\subsection{Dataset}

We used the polarization denoising and demosaicking dataset published in~\cite{11084558} containing 40 scenes. Among the provided three noise-level conditions, we used the highest noise-level condition (High) to evaluate our proposed method under a severe noise condition. 
The dataset was split into 30 scenes for training, 5 scenes for validation, and the rest 5 scenes for testing.

\subsection{Compared Methods}

For the joint task of polarization DM and DN, existing studies mainly focus on a single processing module instead of the complete pipeline. Thus, we compared our proposed method with the methods in three categories: (1) DM Only: This category performs only DM. We adopted one signal-processing-based DM method (IGRI-2)~\cite{9583277} and one deep-learning-based method (TCPDNet)~\cite{nguyen2022two}, which is the backbone for our color DM. (2) DM-to-DN: As a straightforward pipeline, we combined IGRI-2 and BM3D~\cite{dabov2007image} as a signal-processing-based method, and combined TCPDNet and the same U-Net DN architecture as ours as a deep-learning-based method, as shown in Fig.~\ref{fig:pipeline}(a). (3) DN-to-DM: As a signal-processing-based method, we compared the baseline method proposed in~\cite{11084558}, which combines PFCD~\cite{7351714} for DN and IGRI-2 for DM. As a deep-learning-based method, we applied the pipeline shown in Fig.~\ref{fig:pipeline}(b), where we adopted the same U-Net architecture as ours for DN and TCPDNet for DM. To ensure a fair comparison, all deep-learning-based methods are trained on the same dataset. 

\subsection{Results}

Table~\ref{tab:tokyo_results} shows the quantitative evaluation results using peak signal-to-noise ratio (PSNR), structural similarity index measure (SSIM)~\cite{wang2004image}, and mean angular error (MAE). MAE (lower is better) is exclusively used for the evaluation of angle-of-polarization~(AoP), while PSNR and SSIM are applied to all remaining components, including a Stokes parameter ($S_0$) and degree-of-polarization~(DoP). As demonstrated in Table~\ref{tab:tokyo_results}, our proposed method consistently outperforms existing methods across all metrics in the joint task of DM and DN. In particular, our method achieves notable improvements in PSNR and SSIM due to its effective fusion architecture rather than relying on sequential applications of the two tasks. A visual comparison on one of the test scenes is provided in Fig.~\ref{fig:vis}, where we can confirm that our proposed method produces the closest result to the ground truth (GT), especially for the AoP-DoP visualization. The signal-processing-based methods (the methods in the top row) generally produce inferior results, compared with the deep-learning-based methods~(the methods in the bottom row). Importantly, our method significantly outperforms the baseline method proposed in~\cite{11084558} (PFCD$\rightarrow$IGRI-2), which is a sole framework for the joint task of DN and DM for CPFA sensors in the past literature. More visual comparisons can be seen in the supplementary material.

\subsection{Ablation Study}

To validate the effectiveness of our proposed fusion module, we conducted an ablation study. We compared the performance of the complete model (Ours) and the version without the fusion module (Ours w/o Fusion).
The results are shown in Table~\ref{tab:ablation}, where the same evaluation metrics as Table~\ref{tab:tokyo_results} are used. From the results, it can be seen that the fusion module plays a meaningful role in improving model performance. After removing the fusion module, there are decreases in the numerical performance in all the evaluated metrics, which demonstrates the effectiveness of our fusion module.

\begin{table}[t!]
\centering
\caption{Ablation study.}
\vspace{-1mm}
\label{tab:ablation}
\renewcommand{\arraystretch}{1.2}
\setlength{\tabcolsep}{6pt}
\begin{tabular}{lccc}
\toprule
\textbf{Method} & $S_0$ & DoP & AoP \\
\midrule
Ours w/o Fusion & 36.56/0.9427 & 34.21/0.8052 & 42.24 \\
Ours & \textbf{36.79/0.9468} & \textbf{34.50/0.8166} & \textbf{41.47} \\
\bottomrule
\end{tabular}
\end{table}

\section{Conclusion}

In this paper, we have presented CPDDNet, a unified framework of joint DN and DM for CPFA sensors. By integrating a DN-to-DM design with GFMs, our CPDDNet preserves high-frequency structure while effectively reducing noise. Experimental results show that CPDDNet consistently improves visual quality and polarization parameter accuracy compared to existing approaches. As one of the limitations, the training of CPDDNet is currently performed under a fixed noise level. Thus, our future work includes generalizing our CPDDNet to a more diverse range of noise levels using a single model.

\newpage
\bibliographystyle{IEEEbib}
\bibliography{refs}

\end{document}


%
\maketitle

\section{Model Complexity}
\label{sec:model_complexity}

Table~\ref{tab:model_complexity} reports the parameter count and FLOPs of the compared pipelines under a batch size of 1 and an input RAW resolution of $768 \times 1024$. The FLOPs are measured on the real forward path. Our CPDDNet has the largest parameter count because it combines a raw-domain denoising module with the fusion-based demosaicking network. However, its computational cost remains lower than the DM$\rightarrow$DN pipeline.

The main reason is that FLOPs are determined not only by the number of parameters, but also by the spatial resolution at which each module operates. In CPDDNet, the denoising module is performed in the channel-separated RAW domain with a reduced spatial size of $16 \times 192 \times 256$, which is much cheaper than applying denoising after demosaicking at full-resolution RGB/polarization features. As a result, although our method uses more parameters overall, it still requires fewer FLOPs than the DM$\rightarrow$DN pipeline.

\begin{table}[H]
\centering
\caption{Model complexity comparison. FLOPs are measured with input size $1 \times 1 \times 768 \times 1024$.}
\vspace{2mm}
\label{tab:model_complexity}
\small
\begin{tabular}{lcc}
\toprule
Method & Parameters & FLOPs \\
\midrule
DM & 34.522M & 3.854T \\
DN$\rightarrow$DM & 65.562M & 3.924T \\
DM$\rightarrow$DN & 42.225M & 4.707T \\
CPDDNet (Ours) & 68.647M & 4.415T \\
\bottomrule
\end{tabular}
\end{table}

\section{Scene-wise Results}
\label{sec:polar_results}

\begin{figure*}[t!]
\centering
\includegraphics[width=1\textwidth]{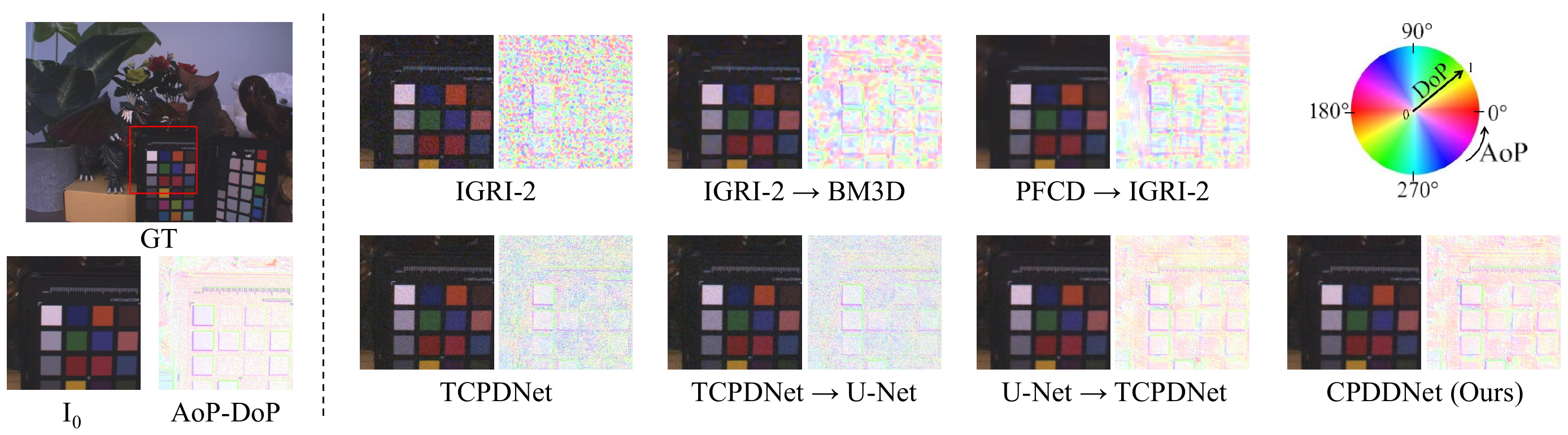}\\ \vspace{-3mm}
\caption{Scene 36.}
\label{fig:scene36}
\end{figure*}

\begin{figure*}[t!]
\centering
\includegraphics[width=1\textwidth]{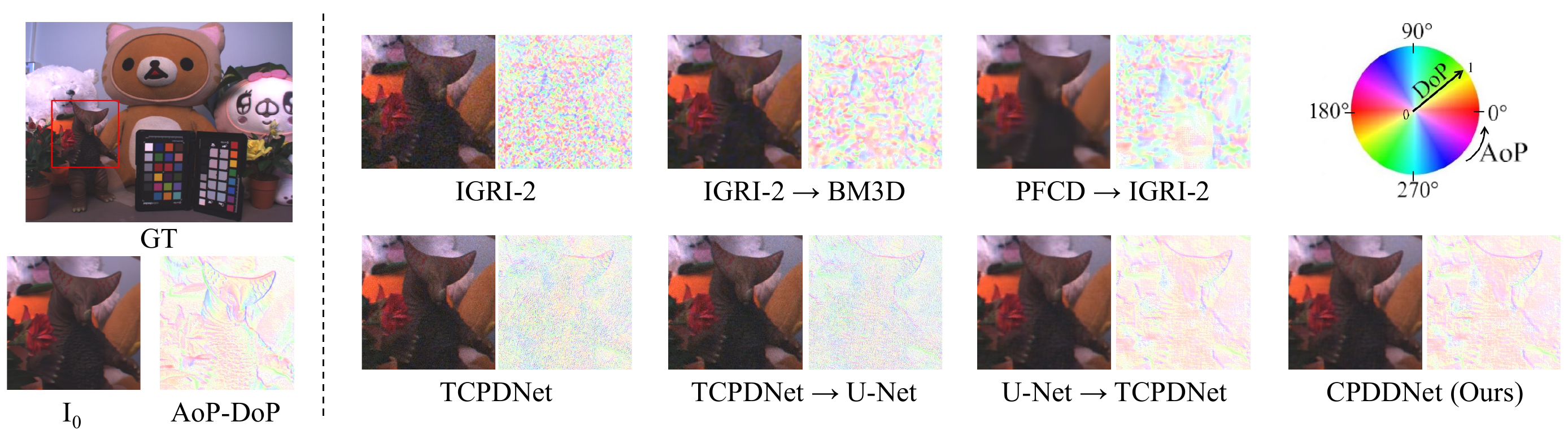}\\ \vspace{-3mm}
\caption{Scene 37.}
\label{fig:scene37}
\end{figure*}

\begin{figure*}[t!]
\centering
\includegraphics[width=1\textwidth]{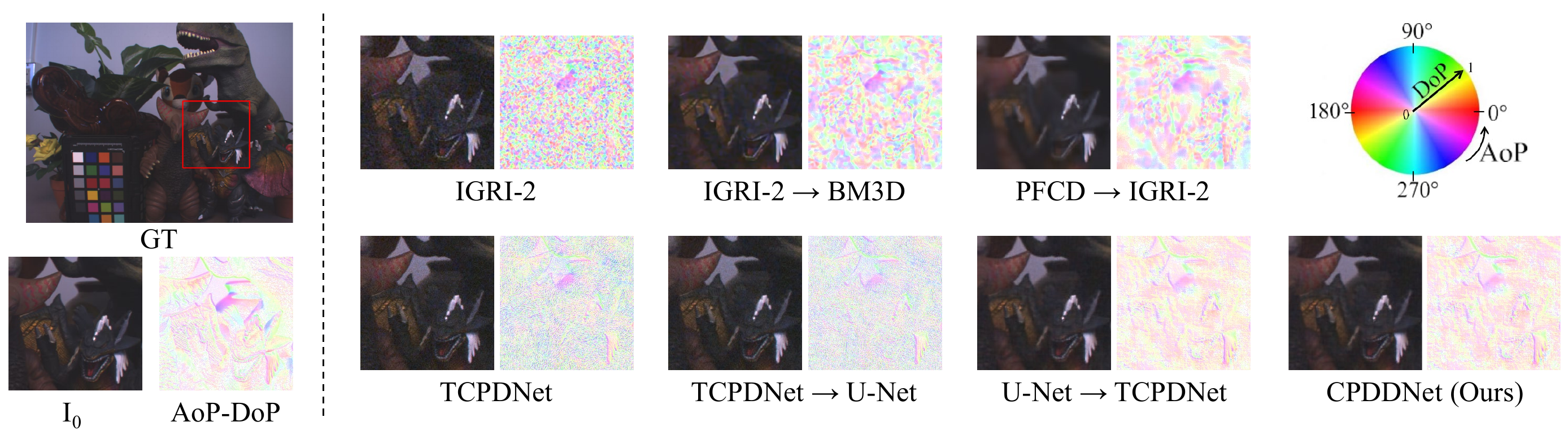}\\ \vspace{-3mm}
\caption{Scene 38.}
\label{fig:scene38}
\end{figure*}

\begin{figure*}[t!]
\centering
\includegraphics[width=1\textwidth]{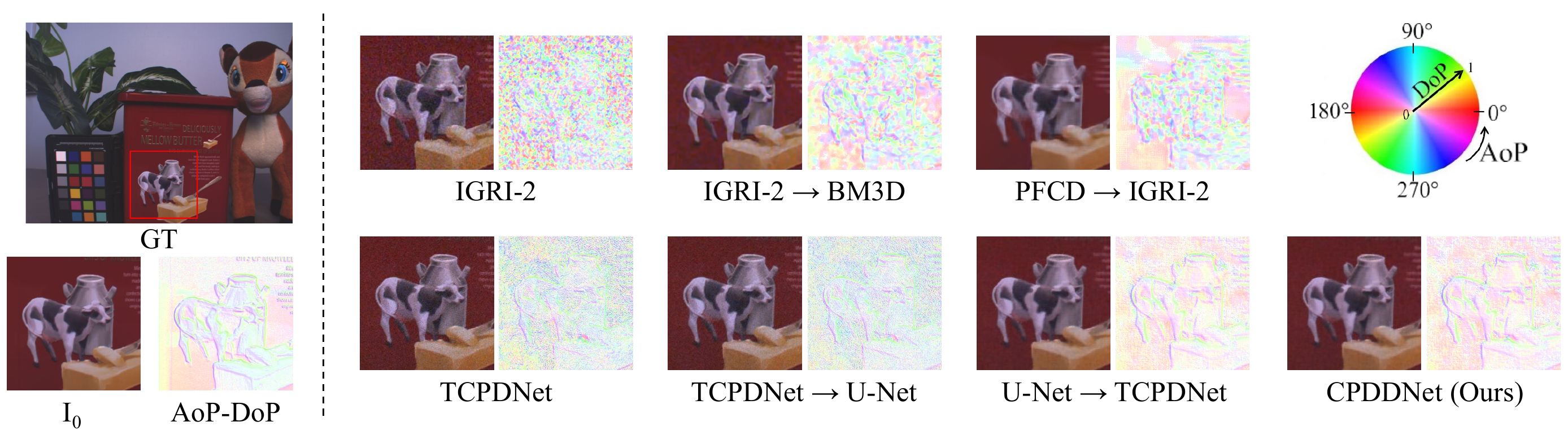}\\ \vspace{-3mm}
\caption{Scene 39.}
\label{fig:scene39}
\end{figure*}

\begin{figure*}[t!]
\centering
\includegraphics[width=1\textwidth]{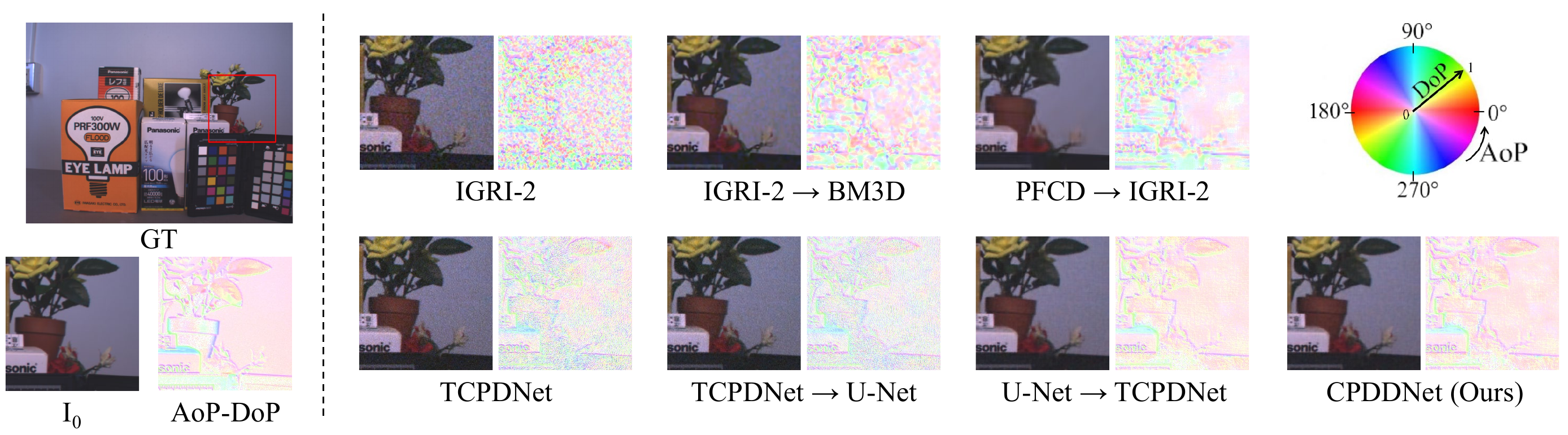}\\ \vspace{-3mm}
\caption{Scene 40.}
\label{fig:scene40}
\end{figure*}

This section includes the visual comparisons for all five tested scenes, which cannot be included in the main paper due to limited space. Figures~\ref{fig:scene36} to \ref{fig:scene40} from the next page show the results for the five scenes (Scenes 36--40), where our proposed method consistently produces the closest reconstructions to the ground truths, especially in the AoP--DoP visualization.

To complement the visual comparisons, we further summarize the scene-wise quantitative ranking over all evaluated methods. For each scene, we compare five methods over 17 metrics, including CPSNR and SSIM on the four polarization angles, the three Stokes components, and DoP, together with the AngleError of AoP. We assign the best metric credit to the top method for each metric. Under this criterion, our CPDDNet ranks the first in all five scenes, obtaining 17/17, 17/17, 16/17, 15/17, and 14/17 best-metric credits on Scenes 36--40, respectively.

\begin{table}[H]
\centering
\caption{Scene-wise quantitative summary. ``Credit'' denotes the accumulated best-metric credit over the 17 evaluated metrics in each scene.}
\vspace{2mm}
\label{tab:scene_summary}
\small
\begin{tabular}{lcc}
\toprule
Scene & Top Method & Credit \\
\midrule
Scene 36 & CPDDNet (Ours) & 17/17 \\
Scene 37 & CPDDNet (Ours) & 17/17 \\
Scene 38 & CPDDNet (Ours) & 16/17 \\
Scene 39 & CPDDNet (Ours) & 15/17 \\
Scene 40 & CPDDNet (Ours) & 14/17 \\
\bottomrule
\end{tabular}
\end{table}

\clearpage

